# Developing Social Robots with Empathetic Non-Verbal Cues Using Large Language Models

Yoon Kyung Lee[1], Yoonwon Jung[1], Gyuyi Kang[1], and Sowon Hahn[1]
Seoul National University

*Abstract*—We propose augmenting social robots' empathetic capacities by integrating non-verbal cues. Our primary contribution encompasses designing and labeling four types of empathetic non-verbal cues (SAFE: Speech, Action (gesture), Facial expression, and Emotion) in a social robot, employing a Large Language Model (LLM). We developed an LLM-based conversational system for a social robot and assessed the alignment of the social cues with those defined by human counselors. Our preliminary results reveal distinct patterns in LLM-based responses, including a preference for calm and positive social emotions ('joy', 'lively'), and frequent nodding gestures. Despite these patterns, our innovative approach has facilitated the development of a social robot capable of context-aware and more authentic interactions. Our work establishes a foundation for future human-robot interactions, emphasizing the pivotal role of verbal and non-verbal cues in constructing social and empathetic robots.

*Index Terms* - Empathetic Communication, Large Language Models (LLM), Human-Robot Interaction (HRI), Alignment, Context-aware Interactions, Empathy in AI, Social Intelligence, Robot Counseling, AI Robot

## I. Introduction

In-depth communication requires active listening and the use of various non-verbal social cues (e.g., gestures). This gets more challenging for robots, and they need to combine spoken communication with non-verbal social cues, such as body language and speech tone, to gain user trust. Although a significant amount of human-robot interaction (HRI) research has explored the impact of non-verbal cues on improving communication with humans, the question of how to incorporate these into robots' cognitive systems is less examined, especially within the realms of natural language processing (NLP) and artificial intelligence (AI).

The advancement of the large language model (LLM) in NLP has led to enhanced human-level understanding and text generation. However, despite the proliferation of text-based conversational agents based on these NLP technologies, further research is needed to align non-linguistic cues for real-time empathetic communication between humans and AI (robots).

Furthermore, a social robot must establish trust so that humans feel comfortable sharing their feelings and thoughts. This necessitates robots demonstrating a comprehensive understanding and empathy toward human clients' needs. This gap leads to our central question: Can we enhance empathy and active listening in such scenarios by developing a unified cognitive system that incorporates a large language model and proposes suitable non-behavioral cues in each situation?

[1]Yoon Kyung Lee, Yoonwon Jung, Gyuyi Kang, Sowon Hahn are with the Human Factors Psychology Lab, Department of Psychology, Seoul National University, 08826, Seoul, Korea. `{yoonlee78, ywjung, qe1995, swhahn}@snu.ac.kr`

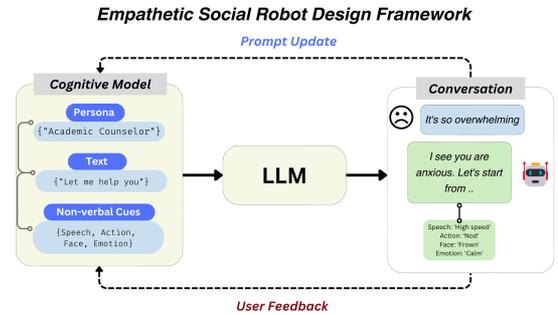

Fig. 1: Flow of developing a cognitive model of social robot for empathetic communication using a large language model. The counselor model then generates an empathetic response with the following non-verbal cues: speech tone, action, facial expression, and emotion. We then tested multiple times with an actual user (who role-plays a client and counselor) in a simulated academic counseling session.

To achieve this, we developed a prompt-based conversational robot that generates non-verbal cues using LLM and evaluated whether they align with human counselors in simulated online counseling sessions. Our approach focuses on the counselor's messages by matching them with the corresponding Speech tone, Action (gesture), Facial expression, and Emotion (SAFE). This method will foster more authentic and context-aware interactions, underscoring the importance of both linguistic and paralinguistic cues. Our study will suggest a novel way to enhance human-robot communication, leading to more effective communication.

## II. Related Work

### A. Empathy in Human-Robot Interaction

Empathy is crucial for social interactions, and it involves understanding and responding to others' emotions and mental states [1]. Psychological theories define empathy as the ability to perceive and respond to others' mental states and emotions, involving affective resonance (sympathy or compassion) and theory-of-mind (perspective-taking) [2]. Different aspects of empathy influence social relationships to varying degrees, promoting understanding and prosocial behaviors [3], [4]. This capacity for affective resonance (to be able to recognize one's emotion and reflect it in one's mental state and feeling) and perspective-taking is increasingly vital in HRI, particularly with its implications for rapport building and informative communications such as in robot tutoring, robot counseling, or human-robot collaboration [5], [6]. Despite advancements in emotion recognition [7], [8] and robotic empathy (e.g., [9], [10]), there remains an under-representation of non-verbal cues in empathy-based HRI that align with humans' preferences.

*B. The Role of Non-Verbal Cues in Empathetic Responses*

Non-verbal cues are fundamental to deciphering intent and emotion in human-human communication. Many studies in psychology or counseling have narrowed down many non-verbal cues that facilitate accurate emotion recognition (FACS; [11]) and active listening in therapy [12]. However, most of these studies are done with the human population and lack a comprehensive integration of multiple non-verbal cues, especially in human-robot communication. Moreover, these studies have not developed robots' mental models and courses of action most desirable in social interactions.

*C. Large Language Models for Context-aware Communication*

LLMs are transformer-based probabilistic language models trained on massive amounts of text corpus [13], [14]. For example, GPT-3 (Generative Pre-trained Transformer models) was trained on 45 terabytes of text data and around 175 billion parameters to understand and generate contextually appropriate, human-like text [13].

LLMs have shown ground-breaking performance and unprecedented ability in language understanding and generation. These models are notable for their zero-shot and one-shot learning capabilities. Zero-shot learning allows a model to handle tasks not explicitly encountered during training, inferring the expected response from the prompt [13]. In contrast, one-shot learning involves task performance using a single or a few examples in the prompt. This has advanced in the technique of 'prompt engineering,' strategically designing the input or 'prompt' to enhance model performance and to make the output align with human preference [15]. Applied to conversational agents, this technique bolsters the generation of precise, context-aware responses, enabling more natural communication [15], [16]. Further enhancements have been achieved by adding features such as consistent personas [17], long-term memory [18], and psychological attributes.

*D. The Present Study*

The present study focuses on utilizing generative AI and LLM to examine the alignment of empathetic responses in AI-based social robots with human counterparts, with an emphasis on non-verbal cues. We propose a research framework for testing robots' communication skills and set a standard for evaluating robots' empathetic responses. Additionally, we suggest a practical experimental design based on LLM prompts that enable natural interactions between humans and robots, calibrated to align with human preferences and goals.

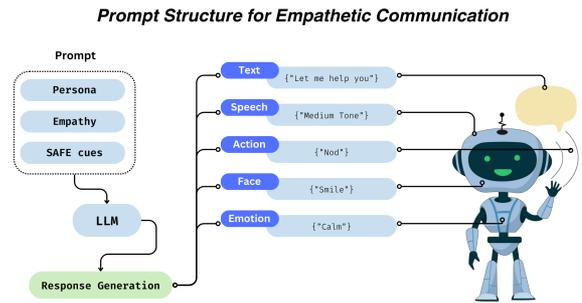

Fig. 2: Prompt structure for aligning non-verbal cues for a social robot in counseling scenario.

## III. DESIGN

Fig.1 depicts the flow of the counseling social robot demonstration.

*1) System Setup:* We used GPT-3.5 API from OpenAI [1] for system setup. The system takes a user query as input and returns a dictionary containing the status and response from the API. Several parameters are specified for the large language model. The model('text-davinci-003') temperature is set to 0.9, controlling the randomness of the generated text (with 1 being the most random). The maximum token limit was set to 200 to control response length. The *top p* parameter is set to 1 for nucleus sampling to reduce the randomness of the output (The frequency penalty = 0 and the presence penalty = 0.6). The stop tokens include 'Human:' and 'AI:' indicating the API to stop generating additional text when it encounters these tokens.

*2) Prompt Design*: Fig. 2 shows the structure of the prompt which includes the cognitive model of the social robot (persona, empathy, and non-verbal cues). The prompt is formed by concatenating 'instructions,' 'format,' and an 'example'.

*3) Empathetic Non-Verbal Cue Definition*: Table I shows non-verbal cues detectable in everyday communication: Speech tone, Action (gesture), Facial expression, and Emotion (SAFE). The cues were defined based on previous studies from psychology and counseling [11], [19], [20], [21]. We additionally created multiple options that represent each cue (e.g., 'High and slow tone' for Speech tone) and labeled numerical labels per option (Table II).

For the 'Speech' cues, we grounded our options on previous research on how speech-related cues are perceived to be associated with the other characteristics of an artificial agent (i.e., personality) [19]. For the 'Emotion' cues, we referred to the basic emotion category and then modified the options by adding more social emotions (e.g., joy, worry) that were more plausible in counseling. We then developed a scoring system where each cue category can be evaluated based on whether they align with the ground truth of humans' responses to the same question. If the robots' response to the same question aligns with the one defined by human ground truth, we evaluated it as 1 and 0 for the non-aligned case. The criteria for scoring are as follows: Human evaluators defined the most appropriate non-verbal cues for each potential user input message and counselor's responding message. We compared this predefined ground truth (human responses) to the robot's responses generated by LLM to calculate the alignment score. For scoring, 1 (aligns with the human counselor's response) or 0 (does not align) was assigned to each non-verbal cue category.

---

[1] https://openai.com/

TABLE I. OUR SAFE NON-VERBAL CUES

| Non-verbal cues | Options |
|---|---|
| **S**peech (7) | 1: High and fast speech, 2: High and medium pace speech, 3: Low and slow speech, 4: Low and moderately fast speech, 5: Fast speech in neutral tones, 6: Medium-paced speech in neutral tones, 7: Slow speech in neutral tones |
| **A**ction (7) | 1: Turn your head towards the speaker, 2: Shake head, 3: Put your hands on your shoulders, 4: Raise one hand diagonally upward, 5: Nod, 6: Interlock hands and place them on the table, 7: Eye Contact |
| **F**acial expression (10) | 1: Frown, 2: Light Smile, 3: Pout, 4: No expression, 5: Bright Smile, 6: Raise eyebrows, 7: Grin, 8: Lower the tips of your eyebrows, 9: Jaw drop, 10: Widened Eyes |
| **E**motion (10) | 1: Joy, 2: Lively, 3: Sad, 4: Surprised, 5: Angry, 6: Worry, 7: Calm, 8: Indifferent, 9: No emotion, 10: Disgust |

TABLE II. COMPARISON OF DIALOGUES BETWEEN HUMAN-HUMAN AND HUMAN-ROBOT

| Speaker | *Dialogue* |
|---|---|
| Human Client | "I am too nervous for the upcoming internship interview" |
| Human Counselor | *"Don't worry! Shall we come up with a specific plan to prepare for the interview?"* **S**peech: High and fast speech (opt. 1) **A**ction: Eye contact (opt. 7) **F**acial expression: Frown (opt. 1) **E**motion: Worry (opt. 6) |
| Robot Counselor | *"You must be feeling anxious. Let's devise a solid preparation strategy for your interview."* **S**peech: Medium-paced speech in neutral tones (opt. 6) **A**ction: Eye contact (opt. 7) **F**acial expression: Neutral expression (opt. 8) **E**motion: Worry (opt. 6) |

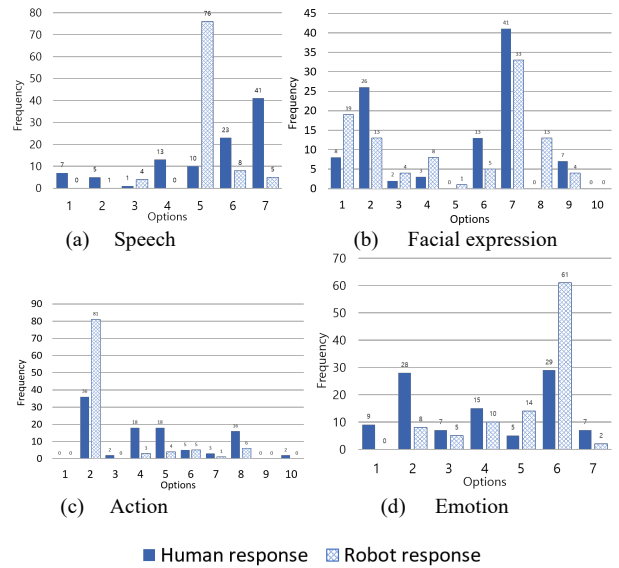

Fig. 3: Frequency distribution of human and robot responses by non-verbal cues.

## IV. USER EVALUATION

The full descriptive statistics of the alignment score are in Table III. The test data were collected from two participants, and a total of 100 question-answer pair data points were analyzed. We evaluated the alignment of the messages and non-verbal cues of a human and robot response generated to the same user input question. We computed the alignment score for each modality (e.g., a match in speech cue for human response and robot response). The mean alignment score across nonverbal cues was 0.25, equivalent to 24.75% accuracy ($SD$=0.24). The alignment score and accuracy were highest for emotion ($M_{score}$=0.32, accuracy=32%, $SD$=0.47) and comparably high for facial expression (face) ($M_{score}$=0.31, accuracy=31%, $SD$=0.46). The alignment score and accuracy were lowest for action ($M_{score}$=0.10, accuracy=10%, $SD$=0.30).

Fig. 3 presents the distributions of non-verbal cues in human and robot responses. In the case of 'Speech' cues, 'Medium-paced speech in neutral tones' (option 6; MSNT) was the most frequently observed in both human and robot responses. While the robot predominantly used the MSNT to convey empathy (61% of all responses), human responses displayed a more balanced frequency distribution with MSNT at 29 % followed by 'High and medium-paced speech' (option 2) at 28 %.

In terms of 'Action' cues, the robot responses showed a strong preference for 'Nod' (option 5), representing 76% of total responses. Conversely, 'Eye contact' (option 7) was most frequently observed in human response, with comparatively more even distributions. 'Eye contact' (option 7) was followed by 'Interlock hands and place them on the table' (option 6), each constituting 41 % and 23 %, respectively.

Regarding the 'Facial expression' cues, the most common response option for humans and robots was 'Light smile' (option 2). Conversely, human responses were more evenly distributed: 'Light smile' was followed by 'No expression' (option 4) and 'Bright smile' (option 5), each accounting for 18%.

Lastly, for 'Emotion' cues, humans and robots' most frequently selected response option was 'Calm' (option 7), comprising 41 % and 33 %, respectively. The second most frequent option among human responses was 'Lively' (option

TABLE III. NON-VERBAL CUE ALIGNMENT BETWEEN HUMAN AND ROBOT RESPONSES

|  | Speech | Action | Face | Emotion | Total |
|---|---|---|---|---|---|
| Score | 0.26 | 0.10 | 0.31 | 0.32 | 0.25 |
| SD | 0.44 | 0.30 | 0.46 | 0.47 | 0.22 |

2) and 'Joy' (option 1) among robot responses, each constituting 26 % and 19 %.

## V. CONCLUSION

In this exploratory study, we examined the role of LLM in creating a mental model necessary for an empathetic social robot capable of informative counseling and whether the non-verbal signals necessary for conveying empathy align well with human responses.

While our results are still preliminary, we discovered the following patterns. First, even for the same question, humans prefer to use a variety of non-verbal cues even if the word expression or context changes slightly. Second, the empathetic nuance generated by LLM in one-shot is significantly higher quality than typical human responses. Still, upon further examination, the responses often read too general or, from the perspective of a direct conversation participant, lacking interest or providing superficial short answers without actually addressing or counter-questioning the emotion and situation of the client (client's question of "I want to quit my job.." with robot response as "That's a great idea ...", and with the human response "Oh no, what happened? can you tell me more about it?".)

However, given the context of the conversation set in our study assumed a more professional and objective counseling situation (i.e., academic counseling), we identified some intriguing patterns. First, as a counseling robot tuned for more informative conversations rather than merely showing sympathy, the robot frequently suggested 'calm' emotions. It also showed a tendency to prefer to default positive social emotions ('Joy,' 'Lively'). This tendency also aligned with human-defined ground truth, suggesting that humans expect the robot to be calm, informative, or cheerful in such counseling contexts. Secondly, 'Action' cues corresponding to the social robot's gestures showed the lowest alignment. This was because the robot mostly preferred to express empathy with 'Nod', which is not always the most appropriate gesture from the human perspective.

Nonetheless, our findings and research framework enable social robots to demonstrate more empathetic and informative communication with humans. We developed LLM-based mental models for social robots, enabling them to utilize non-verbal cues accompanying their communication. Furthermore, we devised evaluation criteria to refine the linguistic and non-verbal cues required for natural human-robot interaction. This innovative approach paves the way for developing social robots capable of empathetic conversation and contextual understanding.